\documentclass[conference]{IEEEtran}
\IEEEoverridecommandlockouts
\usepackage{cite}
\usepackage{amsmath,amssymb,amsfonts}
\usepackage{algorithmic}
\usepackage{graphicx}
\usepackage{textcomp}
\usepackage{xcolor}
\def\BibTeX{{\rm B\kern-.05em{\sc i\kern-.025em b}\kern-.08em
    T\kern-.1667em\lower.7ex\hbox{E}\kern-.125emX}}

\begin{document}

\title{Cybersecurity Assessment of Smart Grid Exposure Using a Machine Learning Based Approach}

\author{\IEEEauthorblockN{Mofe O. Jeje}
\IEEEauthorblockA{\textit{Department of Computer Science} \\
\textit{School of Electrical Engineering and Computer Science} \\
\textit{University of North Dakota}\\
Grand Folks, North Dakoka, United States \\
mofe.jeje@ndus.edu}}

\maketitle

\begin{abstract}
Given that disturbances to the stable and normal operation of power systems have grown phenomenally,  particularly in terms of unauthorized access to confidential and critical data, injection of malicious software, and exploitation of security vulnerabilities in a poorly patched software among others; then developing, as a countermeasure, an assessment solutions with machine learning capabilities to match up in real-time, with the growth and fast pace of these cyber attacks, is not only critical to the security, reliability and safe operation of power system, but also germane to guaranteeing advanced monitoring and efficient threat detection. Using the Mississippi State University and Oak Ridge National Laboratory dataset, the study used an XGB Classifier modeling approach in machine learning to diagnose and assess power system disturbances, in terms of Attack Events, Natural Events and No-Events. As test results show, the model, in all the three sub-datasets, generally demonstrates good performance on all metrics, as it relates to accurately identifying and classifying all the three power system events.
\end{abstract}


\section{Introduction}
\vspace{-10pt}
Today’s electrical power grid which has become integral and an indispensable part of modern society, has been largely threatened by growing risks of extreme events from, not only the natural disasters, but also by malicious attacks from cyber adversaries [1], targeting several key power system operational functions, such as automatic generation control (AGC), state estimation (SE), and energy management systems among many others (EMS). 

While threats from extreme events, in terms of natural disasters have been largely caused by climate change being somewhat induced, in part by inefficient power grid operations along with other engineering activities around the world; the threat from malicious attacks which falls within the scope of this study, is occasioned, on the other hand, by integration and deployment of the ever-changing Information and Communication Technologies (ICT) into the operation of electric power grid, with the overall goal of intelligently monitoring, controlling and protecting the power system. 

Smart grid, being an answer to myriads of issues that characterized traditional power transmission and distribution systems, in terms of addressing the intermittent nature of renewable energy, mis-match between generated electricity and demand, energy waste, huge revenue losses, consumer non-participation, and an inefficient routing and dispensation of electricity among others; has, in another stretch, evoked concern and attracted significant attention towards the security of power systems. Even though the introduction of ICT into the power systems makes the power grid networks to become smarter, it also substantially expands current cyber attack surface area, by exposing the power grid to vulnerabilities that increases the risk of malicious cyber attacks from adversaries who may alter the underlying physical systems and processes, that could potentially compromise a country’s national security.

Worse still, while the Supervisory Control and Data Acquisition (SCADA) network which has been traditionally used for monitoring, controlling, and protection of power systems has a recurrent problem of low measurements of refresh rate that limits or inhibits real-time assessment of a wide area network; the use of Synchrophasor networks as a replacement, even though has high-speed digital signal processors, with reporting rates sufficiently high to satisfy a number of real-time power system applications [2], also introduces numerous vulnerabilities that come with the risk of cyber-attacks that could cause catastrophic damages to equipment and create large scale power outages..

While attack sources can vary from those that are state-sponsored, terrorist, amateur attackers unaware, or disgruntled employees, deliberate criminals and many more; the objective of an adversary may not just be on gaining unauthorized information; but could also in theory, ultimately include crippling of power grid, by attacking the state estimator with fake meter data, with an intent to mislead the energy management system (EMS) which collects data at the control center from remote meters, to potentially make erroneous decisions on contingency analysis, dispatch, or even billing.

Given that disturbances to the stable and normal operation of power systems have grown phenomenally,  particularly in terms of unauthorized access to confidential and critical data, injection of malicious software, and exploitation of security vulnerabilities in a poorly patched software among others; then developing, as a countermeasure, an assessment solutions with machine learning capabilities to match up in real-time, with the growth and fast pace of these cyber attacks, is not only critical to the security, reliability and safe operation of power system, but also germane to guaranteeing advanced monitoring and efficient threat detection. The rest of the paper is organized as follow: section 2 reviews related works in the literature; section 3 discusses the research methodology; section 4 covers evaluation and discussion of experimental results; finally, section 5 draws the conclusion and the research future direction.

\section{REVIEW OF LITERATURE}
\vspace{-5pt}
\section *{Applicability of Machine Learning in Detection Techniques}
\vspace{-5pt}
Strategies to detect cyber intrusions are plentiful and endless, as there is a growing set of cyber attack surfaces and vectors that can manipulate the grid towards an intruder’s favor. 
 
One of such strategies is the detection and mitigation of DOS attack in AGC. In addition to Virtual synchrophasor networks (VSNs) serving as a countermeasure for DOS attack in synchrophasor network; a port hopping technique, based on multi-path transmission control protocol is proposed in [42] for synchrophasor networks. While a distributed topology formation algorithm in [3] is proposed to isolate the malicious PMU nodes in a power network, [4] presented a detection technique for manipulated PMU measurements using the equivalent impedance of transmission lines.

Another strategy is the Detection and mitigation of False Data Injection Attack (FDIA). [5] proposes a federated learning-based FDIA detection method that utilizes data from all nodes to collaboratively train a detection model, in addition to combining the Paillier cryptosystem. With a different perspective, [6] also presents weighted least square (WLS) based detection scheme for FDIA attacks which, in line with the works of [7],is used for static state estimation in a distributed system, where the focal impact was power system economy.

While work by [8] argued that, with continuous monitoring of correlation coefficients, some of the data spoofing attacks like data tampering or replay attacks can be easily detected; works by [9], and several works have proposed that, detection and mitigation of GPS spoofing attack technique alone cannot mitigate or correct the PMU measurements. A joint estimation algorithm as espoused in [10] is proposed for identifying spoofing signal location and GPS timing. Furthermore, GPS spoofing attack detection and mitigation based on PMU measurement residual are presented in [11], which also estimates the correction of measurements phase angles. 

Much as the above traditional intrusion detection techniques in power systems are good, they are dogged and constrained by the problem of scalability, improved accuracy, reduced false positive, and issue of real-time detection, - all to which Machine learning, a subset of artificial intelligence, empowers systems to learn from vast amounts of data and identify patterns or anomalies that may signal a cyber threat. For this reason, Machine learning and artificial intelligence techniques are gaining traction, as they are now more proposed and applied in power systems to identify disturbances and detect cyber attacks even through deception. 

To detect energy theft, a common challenge in power systems, [12] uses normal and malicious data of consumer consumption patterns and a consumption pattern-based energy theft detector (CPBETD). This tool, combined with the application of a Support Vector Machine (SVM) anomaly detector, allows the algorithm to use silhouette plots to identify different distributions in the dataset while relying on distribution transformer meters to detect nontechnical loss (NTL) at the transformer level. Also, to find the most promising algorithm which can detect an adversarial intrusion, research work by [13], was done to combine SVM with a variety of machine learning algorithms. A robust spam filtering method is introduced in [14] using a hybrid method for rule-based processing along with back-propagation neural network.

[15] in similar trend, created a machine-learned framework, and refined it with unsupervised feature learning to detect different types of cyber attacks in power systems. In comparison with detectors relying on detailed system information and human expertise, a stacked autoencoder-based unsupervised feature learning was proposed to capture useful and rich patterns hidden in the data with which to recognize a cyber attack and achieve competitive results. 

 To address the issue of precision and sensitivity of detection algorithms as it relates to data integrity attacks, advanced machine learning-based techniques for outlier detectors are presented by [16] as a defence mechanism against the occurrence of such attacks.

Given that detecting intrusions through the entire sector of the power network is challenging, [17] presented a proposal of grouping network buses, and designing filters for detection, as well as isolation of faults that addresses a feasible detection mechanism. In similar vein, a devised algorithm by [18] was implemented to accurately detect and locate faults in power systems, in addition to identifying bad data, using weighted least absolute value (WLAV), which has, the ability to reject bad data, to reduce dimensionality. 

Pushing this further, [11] suggested a data-driven algorithm for online power grid topology change identification with PMUs, where the proposed machine learning algorithm can differentiate the various types of faults in power grids, and the topology switching actions initiated by the system operators or attackers.

In drawing inferences from all the review, it is the thesis of this study that, even though there have been quite a few papers in this area, the problem of cyber attack diagnosis or assessment in power system with the application of machine learning has been dogged by a weak generalization due to the limited data availability. It is against this background that this study embarks to investigate this hypothesis from a significantly increased data perspective using machine learning.

\section{Methodology}
\vspace{-10pt}
\subsection{Data Characteristics}\label{AA}
\vspace{-10pt}
This research study uses an eXplainable Artificial Intelligence (XAI) with respect to SHAP, a Machine learning technique used to explain the decision-making process in Machine Learning models, after the use of a detection model that is capable of accurately detecting power system attacks using the dataset (power system events) provided by Mississippi State University and Oak Ridge National Laboratory. 

The power system events dataset contains 892 features with one label, all totalling 4966 rows × 892 columns. The dataset, in addition to being grouped into binary, three-class, and multiclass. consists of three datasets, each containing 37 power system event scenarios. The scenarios are divided into Natural Events, No Events, and Attack Events. Each scenario includes various types of events such as short-circuit faults, line maintenance, remote tripping command injection (attack), relay setting change (attack), and data injection (attack). Only the three-class data is used in this research, meaning that the provided data is only a part of the entire three-class dataset. Specifically, the three classes in the scope of this study are “No-Events”, “Attack Events”, and “Natural Events”. Also, out of the 892 features, only 127 were extracted and used for this study.

\subsection{Data Preprocessing Procedure }\label{AA}
\vspace{-10pt}
After removing ‘infinte’ data from the dataset, and since XGB Classifier, being what is being used in the study requires only numerical inputs; a label encoding, was carried out on the ‘marker’ label column which is the only column requiring conversion from categorical data to numerical data. Also, before using StandardScaler technique to normalize or transform all the input features into a similar scale with the main purpose of improving performance as well as the training stability of model; the dataset was split into 80 percent for training and 20 percent for testing with the use of the train-test-split function of sklearn module in python.

While afterwards, an XGB Classifier was trained with tuned hyperparameters using the training dataset, it was also, on the other hand, evaluated using the testing dataset to determine the model accuracy, along with a confusion matrix to show its performance.

To further improve on the model, a Feature Importance Analysis technique was used to select an optimal subset of 10 input features from all the power system event features that were previously used. Starting with only the extraction of ‘Attack Events and Natural Events’ sub-dataset, the resulting optimal subset of features were subsequently split into 80 percent training set and 20 percent testing set. The XGB classifier was again trained and evaluated respectively on 80 percent and 20 percent of the resulting optimal subsets, in addition to using confusion matrix to visualize its detection performance. In similar fashion, the extraction of ‘Natural Events and No-Events’ followed by ‘Attack Events and No-Events’ sub-datasets were also trained and evaluated subsequently.

\section{RESULTS AND DISCUSSIONS}
\vspace{-10pt}
As shown in figure 1, 2 and 3 below, the Mean absolute SHAP values are typically displayed as bar plots that rank features by their importance. The main characteristics to examine here are the ordering of features, and the relative magnitudes of the mean absolute SHAP values.

\begin{figure}[htbp]
\centerline{\includegraphics[width=0.75\linewidth]{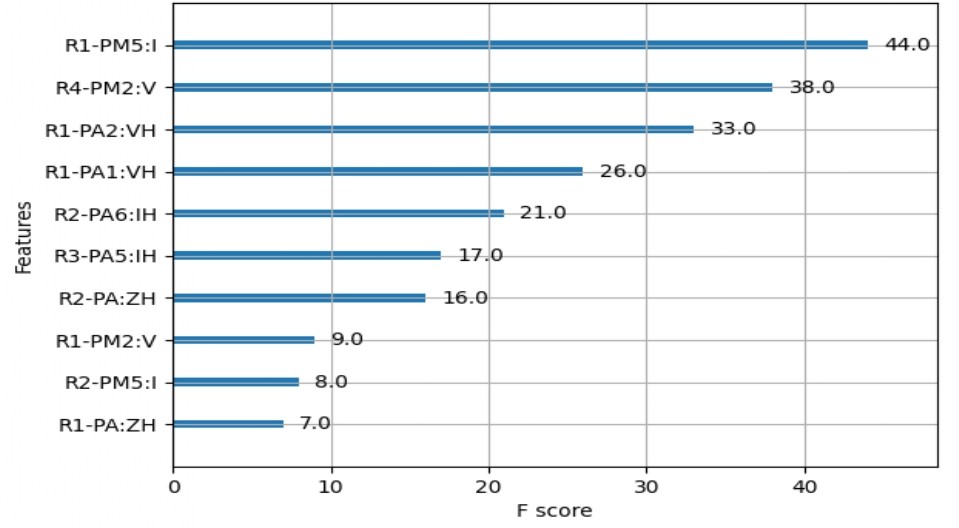}}
\renewcommand{\thefigure}{1a}%
\caption{Summary Plot of Feature Importance for 'Attacks and Natural Events'}
\label{fig}
\end{figure}
In figure 1(a), when considering sub-dataset with only 'Attack Events and Natural Events', we see that, after the input features have been selected by importance, R1-PM5:I is the most influential variable, contributing on average 44.0 to each predicted and classified observations, in terms of Attack Events and Natural Events. By contrast, the least influential variable is R1-PA:ZH which contributes 7.0 to the said observations. 

\begin{figure}[htbp]
\centerline{\includegraphics[width=0.75\linewidth]{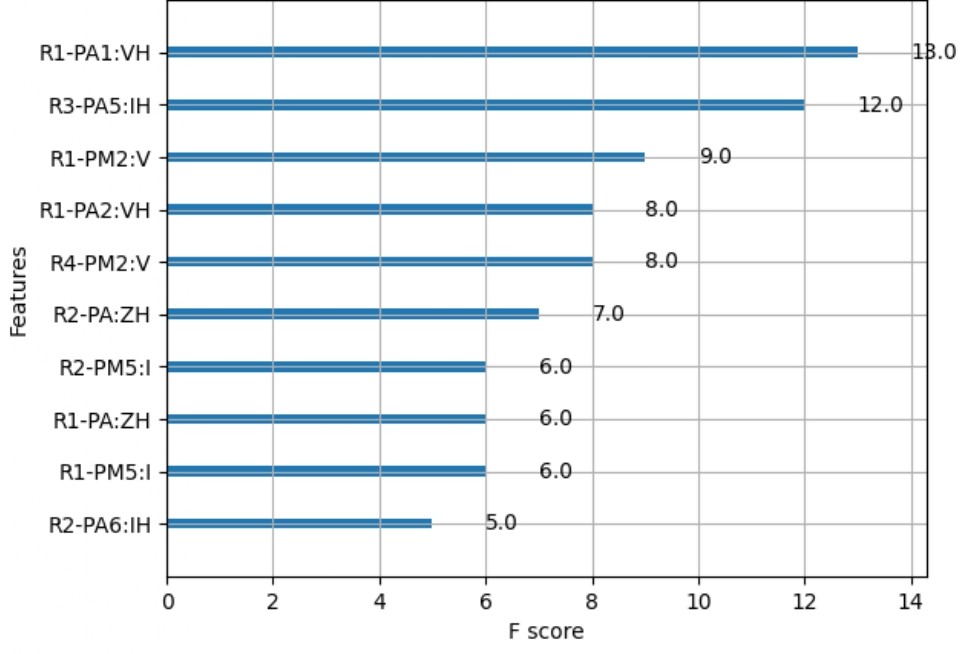}}
\renewcommand{\thefigure}{1(b)}%
\caption{Summary Plot of Feature Importance for ‘No-Events and Natural’}
\label{fig}
\end{figure}

In figure 1(b), however, in considering the sub-dataset with only 'No-Events and Natural Events', we see R1-PA1:VH as the most influential variable, contributing on average 13.0 to each predicted and classified observations. On the other spectrum of the plot, the least influential variable is R2-PA6:IH which contributes 5.0 to the said observations.

\begin{figure}[htbp]
\centerline{\includegraphics[width=0.75\linewidth]{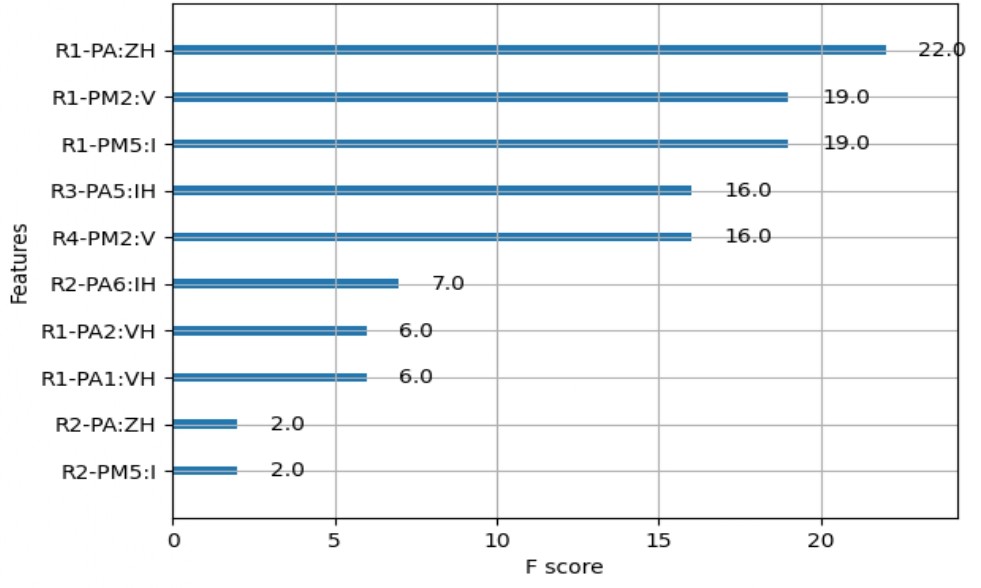}}
\renewcommand{\thefigure}{1(c)}%
\caption{Summary Plot of Feature Importance for ‘Attack and No-Events’}
\label{fig}
\end{figure}

Also in figure 1(c), a consideration of the sub-dataset with only 'Attack Events and No-Events' shows R1-PA:ZH as the most influential variable, contributing on average 22.0 to each predicted and classified observations. On the other spectrum of the plot, the least influential variable is R2-PM5:I which contributes 2.0 to the said observations.
\vspace{-10pt}
\begin{table}[htbp]
\renewcommand{\thetable}{1a}%
\caption{Confusion matrix for Sub-dataset of Attack and Natural}
\begin{center}
\begin{tabular}{|c|c|c|}
\hline
\textbf{Actual}&\multicolumn{2}{|c|}{\textbf{Predicted}} \\
\hline
&\textbf{\textit{Natural Events}}&
\textbf{\textit{Attack Events}} \\
\hline
\textbf Natural Events& 147& 29 \\
\textbf Attack Events& 783 & 0 \\
\hline
\end{tabular}
\label{table:1}
\end{center}
\end{table}

\vspace{-23pt}

\begin{table}[htbp]
\renewcommand{\thetable}{1b}%
\caption{Classification Report for Sub-dataset of Attack and Natural}
\begin{center}
\begin{tabular}{|c|c|c|c|c|c|}
\hline
& \textbf{\textit{Precision}}& \textbf{\textit{Recall}}& \textbf{\textit{F1Score}}& \textbf{\textit{Support}} \\
\hline
\textbf 0& 1.00& 0.16& 0.28& 176 \\
\textbf 1& 0.84& 1.00& 0.91& 783 \\
&&&&\\
\textbf accuracy& & & 0.85& 959 \\
\textbf macro avg& 0.92& 0.58& 0.60& 959 \\
\textbf weighted avg& 0.87& 0.85& 0.80& 959 \\
\hline
\end{tabular}
\label{table:2}
\end{center}
\end{table}

\vspace{-11pt}

From the above table 1(a), the first row is for the ‘Natural Events’. In that row, the model predicted 147 of these correctly and incorrectly predicted 29 of the ‘Natural Events’ as ‘Attack Events’. Looking at the ‘Natural Events’ column; of the 930 ‘Natural Events’ predicted by the model (i.e. the sum of the ‘Natural Events’ column), 147 were correctly predicted as ‘Natural Events’, while 783 of the ‘Natural Events’ were incorrectly predicted as ‘Attack Events’, and likewise in the ‘Attack Events’ column; while 0 was correctly predicted as ‘Attack Events, 29 of the ‘Attack Events’ were incorrectly predicted as ‘Natural Events’.

In table 1(b), while the study under classification report, recorded an accuracy score of 85 percent along with an F1score of 91 percent; it also, in terms of precision and recall respectively achieved 84 percent and 100 percent outcome for correctly classifying and predicting the ‘Attack Events’.

\vspace{-11.2pt}
\begin{table}[htbp]
\renewcommand{\thetable}{2a}%
\caption{Confusion matrix for Sub-dataset of ‘Natural and No-Events’}
\begin{center}
\begin{tabular}{|c|c|c|}
\hline
\textbf{Actual}&\multicolumn{2}{|c|}{\textbf{Predicted}} \\
\hline
&\textbf{\textit{No-Events}}&
\textbf{\textit{Natural Events}} \\
\hline
\textbf No-Events& 1& 37 \\
\textbf Natural Events& 181 & 1 \\
\hline
\end{tabular}
\label{table:1}
\end{center}
\end{table}

\vspace{2em}

\begin{table}[htbp]
\renewcommand{\thetable}{2b}%
\caption{Classification Report for Sub-dataset of Natural and No-Events}
\begin{center}
\begin{tabular}{|c|c|c|c|c|c|}
\hline
& \textbf{\textit{Precision}}& \textbf{\textit{Recall}}& \textbf{\textit{F1Score}}& \textbf{\textit{Support}} \\
\hline
\textbf 0& 0.97& 0.97& 0.97& 38 \\
\textbf 1& 0.99& 0.99& 0.99& 182 \\
&&&&\\
\textbf accuracy& & & 0.99& 220 \\
\textbf macro avg& 0.98& 0.98& 0.98& 220 \\
\textbf weighted avg& 0.99& 0.99& 0.99& 220 \\
\hline
\end{tabular}
\label{table:2}
\end{center}
\end{table}

\vspace{20pt}
\setlength{\textfloatsep}{1pt}
From the above table 2(a), the first row is for the ‘No- Events’. In that row, the model predicted 1 of this correctly and incorrectly predicted 37 of the ‘No-Events’ to be ‘Natural Events’. Looking at the ‘No-Events’ column; of the 182 ‘No-Events’ predicted by the model (i.e. the sum of the ‘No-Events’ column), 1 was correctly predicted as ‘No-Events’, while 181 as ‘No-Events’ was incorrectly predicted as ‘Natural Events’, and likewise in the ‘Natural Events’ column, 1 was correctly predicted as ‘Natural Events’, while 37 of ‘Natural Events’ were incorrectly predicted as ‘No-Events’.

Likewise in table 2(b) above, while the study under classification report, recorded an accuracy score of 99 percent along with an F1score of 99 percent, it also, in terms of precision and recall achieved a 99 percent outcome each for correctly classifying and predicting the ‘Natural Events’.

\begin{table}[htbp]
\renewcommand{\thetable}{3a}%
\caption{Confusion matrix for Sub-dataset of ‘Attack and No-Events’}
\begin{center}
\begin{tabular}{|c|c|c|}
\hline
\textbf{Actual}&\multicolumn{2}{|c|}{\textbf{Predicted}} \\
\hline
&\textbf{\textit{No-Events}}&
\textbf{\textit{Attack Events}} \\
\hline
\textbf No-Events& 24& 16 \\
\textbf Attack Events& 768 & 0 \\
\hline
\end{tabular}
\label{table:1}
\end{center}
\end{table}

\vspace{-2em}

\begin{table}[htbp]
\renewcommand{\thetable}{3b}%
\caption{Classification Report for Sub-dataset of ‘Attack and No-Events’}
\begin{center}
\begin{tabular}{|c|c|c|c|c|c|}
\hline
& \textbf{\textit{Precision}}& \textbf{\textit{Recall}}& \textbf{\textit{F1Score}}& \textbf{\textit{Support}} \\
\hline
\textbf 0& 1.00& 0.40& 0.57& 40 \\
\textbf 1& 0.97& 1.00& 0.98& 768 \\
&&&&\\
\textbf accuracy& & & 0.97& 808 \\
\textbf macro avg& 0.98& 0.70& 0.98& 808 \\
\textbf weighted avg& 0.97& 0.97& 0.96& 808 \\
\hline
\end{tabular}
\label{table:2}
\end{center}
\end{table}

From table 3(a) above, the first row is for the ‘No-Events’. In that row, the model predicted 24 of these correctly and incorrectly predicted 16 of the ‘No-Events’ as ‘Attack Events’. Looking at the ‘No-Events’ column; of the 792 ‘No-Events’ predicted by the model (i.e. the sum of the ‘No-Events’ column), 24 were correctly predicted as ‘No Events’, while 768 of the ‘No-Events’ were incorrectly predicted as ‘Attack Events’, and likewise in the ‘Attack Events’ column, 0 was correctly predicted as ‘Attack Events’, while 16 of the ‘Attack Events’ were incorrectly predicted as ‘No-Events’.

Similarly, in table 3(b) above, while the study under classification report, recorded an accuracy score of 97 percent along with an F1score of 98 percent, it also, in terms of precision and recall respectively achieved 97 percent and 100 percent outcome for correctly classifying and predicting the ‘Attack Events’

\begin{figure}[htbp]
\centerline{\includegraphics[width=0.75\linewidth]{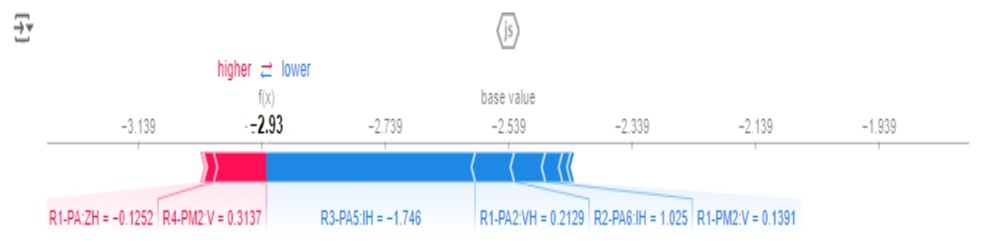}}
\renewcommand{\thefigure}{2}%
\caption{Force Plot}
\label{fig}
\end{figure}
In the above figure 2, each stripe shows the impact of its feature in pushing the value of the target variable farther or closer to the base value. While Red stripes show that their features push the value towards higher values, the Blue stripes on the other hand, show their features push the value towards lower values. The wider a stripe gets, the higher its contribution. The sum of these contributions from all stripes pushes the value of the target variable from the base value to the final, predicted value. 

So, for force plot in figure 2 which involves the 10 optimally selected input features, we see, R1-PA:ZH, and R4-PM2:V values having a positive contribution to the predicted value of -2.93 and a base value 2.54, which is the average value of the target variable across all the records. Out of the 2 variables, R4-PM2:V is the most important variable of record, showing positive contribution with a wider stripe (it has the largest stripe). On the other side of the spectrum, the Blue stripes which push the target variable towards a lower values, have 5 features of which R3-PA5:IH is the widest stripe showing a negative contribution, and R1-PM2:VE is the smallest stripe showing a negative contribution. It is interesting to see, that the value of combined contributions from the 5 Blue stripes with negative contributions, is strong enough to move the predicted value lower than the base value. So, since the total negative contribution (Blue stripes) is larger than the positive contribution (Red stripe), the final value is therefore lower than the base value.

\begin{figure}[htbp]
\centerline{\includegraphics[width=0.75\linewidth]{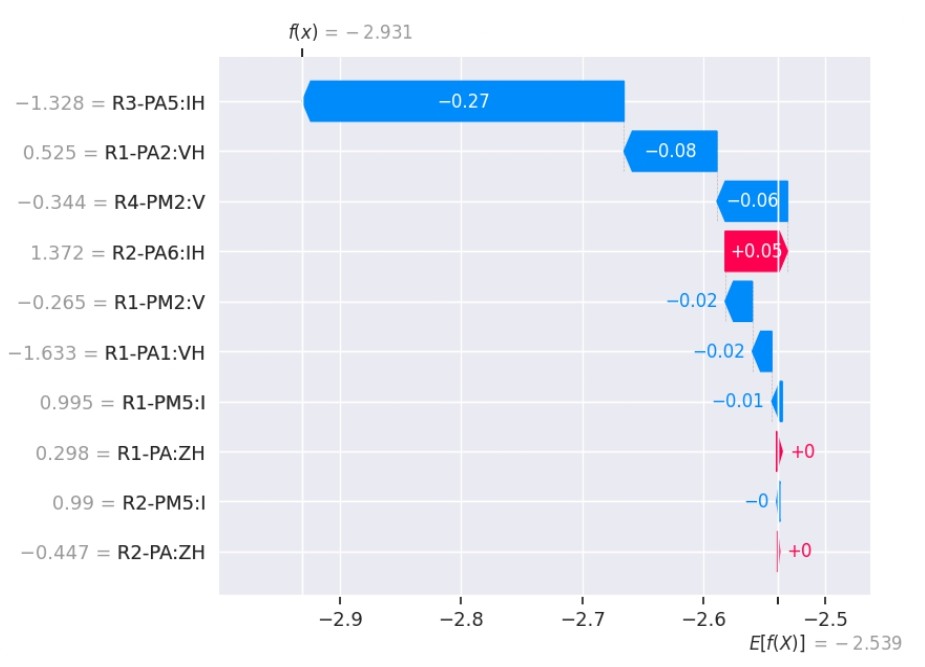}}
\renewcommand{\thefigure}{3}%
\caption{Waterfall Plot}
\label{fig}
\end{figure}

Given the fact that SHAP values of input features will always sum up to the difference between baseline (expected) model output, and the current model output for the prediction being explained, the best way to visualize this is through a waterfall plot that starts at a background prior expectation for power system attacks E[f(X)], and then adds features one at a time until we reach the current model output f(x). Therefore, while waterfall plot in figure 3, shows a background prior expectation for power system attacks with a value of -2.5 that significantly record both negative and positive zero contributions from the  last 3 features at the bottom of the plot; the first 3 features at the top of the plot, on the other hand essentially have a negative contributions which brings the E[f(X)] value of -2.5 to be greater than the current model output f(x) of -2.9.

\begin{figure}[htbp]
\centerline{\includegraphics[width=0.75\linewidth]{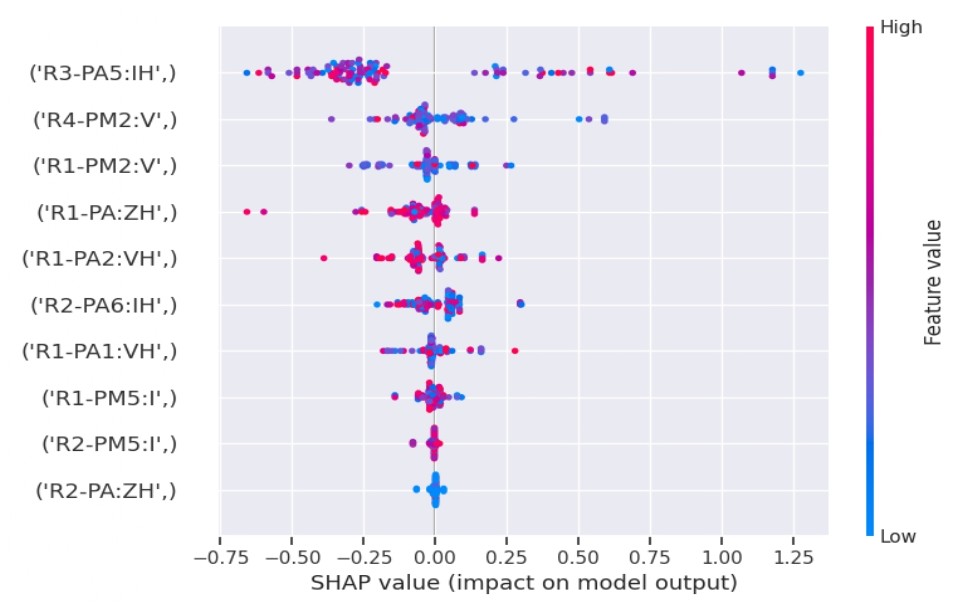}}
\renewcommand{\thefigure}{4}%
\caption{Beeswarm Plots}
\label{fig}
\end{figure}
From the above figure 4, the features are sorted from the most important one to the less important. The higher the value of this feature, the more positive the impact on the target. The lower this value, the more negative the impact of its contribution. So, from figure 4 in the Beewarm plot, while we can see R3-PA5:IH as the most important feature, followed by R4-PM2;V and so on; we also see R2-PA:ZH as the least important feature. 

\begin{figure}[htbp]
\centerline{\includegraphics[width=0.75\linewidth]{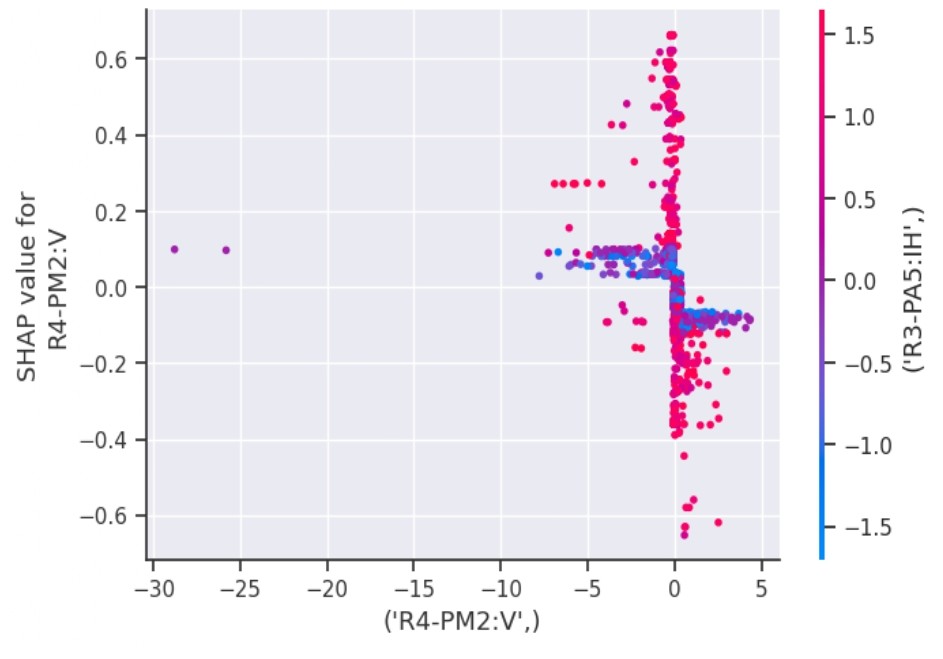}}
\renewcommand{\thefigure}{5(a)}%
\caption{Dependence Plot for R4-PM2:V}
\label{fig}
\end{figure}
A dependence plot is a type of scatter plot that displays how a model's prediction is affected by a specific input feature. On the x-axis, the input features values are displayed, while the y-axis represents the SHAP values associated with each input feature value. In this plot on figure 5(a), the SHAP dependence scatter plot showcases a vertical linear relationship between a power input feature named Voltage Phase Magnitude (R4-PM2:V) and its corresponding SHAP values. The extent of the vertical distribution of the SHAP values indicates how much relative influence the variable has on predictions. From the plot, we observe that the vertical spread of SHAP values which range from roughly -0.4 to 0.6 at a fixed value of zero is due to an interaction effect with of a power input feature named Phase C Voltage Phase Angle (R3-PA5:IH). Therefore, at this region, we see instances where the SHAP values above the y = 0 line is greater than the ones below it, meaning overall that, the Voltage Phase Magnitude, combined with the interaction effect of Phase C Voltage Phase Angle produces a sizeable impact on the predicted SHAP values.

\vspace{-10pt}

\begin{figure}[htbp]
\centerline{\includegraphics[width=0.75\linewidth]{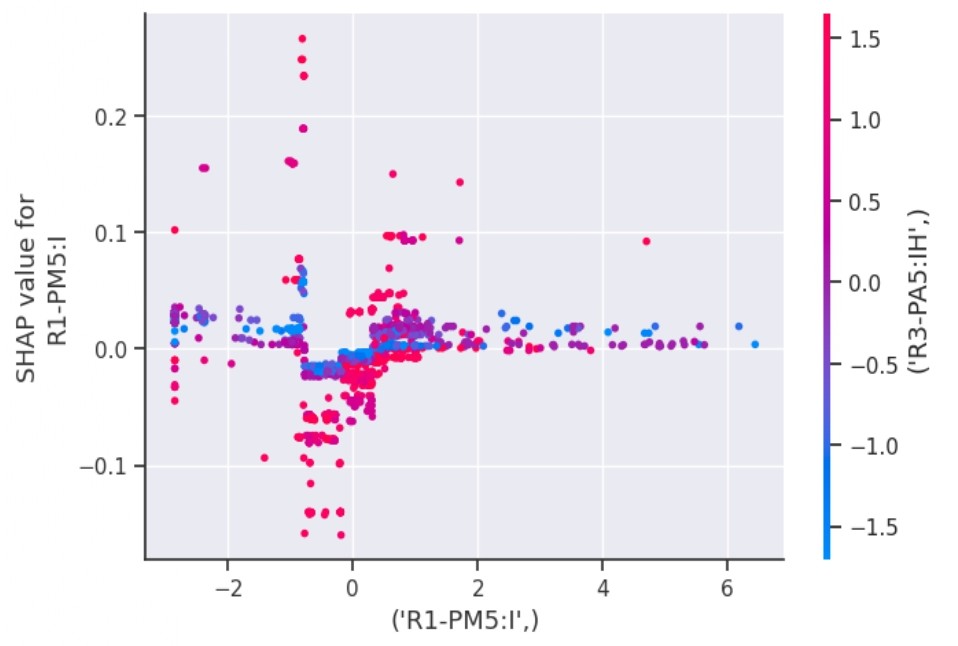}}
\renewcommand{\thefigure}{5b}%
\caption{Dependence Plot for R1-PM5:I}
\label{fig}
\end{figure}
As shown in figure 5(b), examining the dependence plot for R1-PM5:I reveals a horizontal linear relationship between a power input feature named Current Phase Magnitude (R1-PM5:I) and its corresponding SHAP values. From the plot, we observe that the SHAP values below y = 0.0 lines, with the interaction effect of a power feature named Current Phase Angle (R3-PA5:IH) lead to predictions of lower level of the Current Phase Magnitude from 5 to -1, whereas those above it are associated with no visible impact on the R3-PA5:IH. Essentially, the value of R1-PM5:I at which the distribution of SHAP values cross the y = 0.0 line tells us the threshold at which the model switches from predicting lower to higher level of R3-PA5:IH with a corresponding value of  R1-PM5:I at zero.
\vspace{-10pt}

\begin{figure}[htbp]
\centerline{\includegraphics[width=0.75\linewidth]{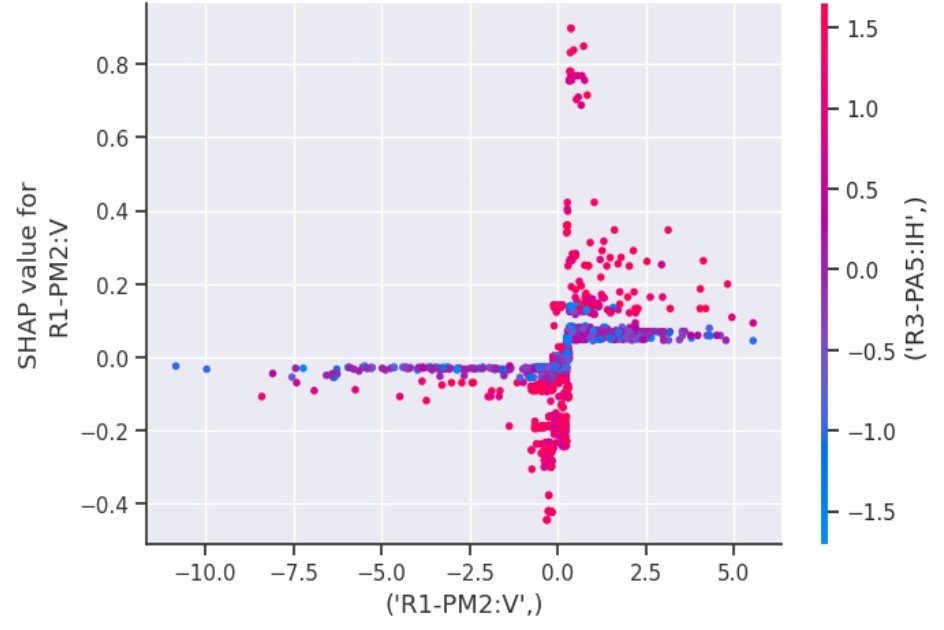}}
\renewcommand{\thefigure}{5(c)}%
\caption{Dependence Plot for R1-PM2:V}
\label{fig}
\end{figure}
In the dependence plot in figure 5(c), the shapes of the distributions of points provide insights into the relationship between the values of a power input feature named Voltage Phase Magnitude (R1-PM2:V) and its corresponding SHAP values. So, for the said dependence plot, we largely see both horizontal and vertical linear relationship across the full range of R1-PM2:V values. From the plot, we observed instances where the SHAP values above the y = 0 line lead to predictions of increases in the values of Voltage Phase Magnitude from 0.0 to 3.0. while the SHAP values below the y = 0 line lead to predictions of decreases in the values of Voltage Phase Magnitude from 0.0 to -3.0. 
\vspace{-10pt}

\begin{figure}[htbp]
\centerline{\includegraphics[width=0.75\linewidth]{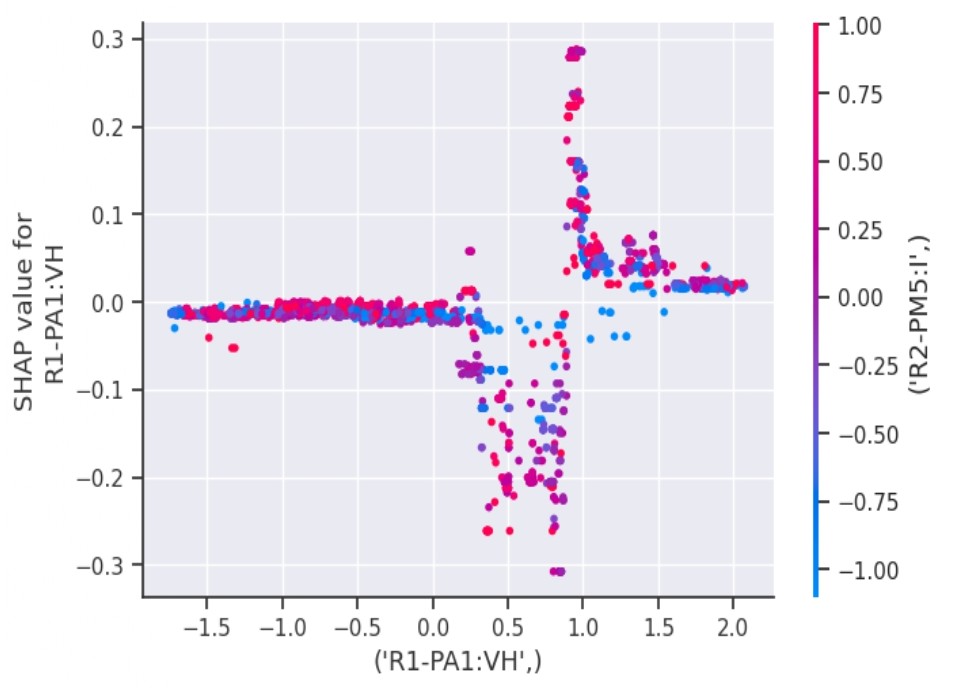}}
\renewcommand{\thefigure}{5(d)}%
\caption{Dependence Plot for R1-PA1:VH}
\label{fig}
\end{figure}

\vspace{10pt}
In examining dependence plot for figure 5(d), it is observed that, while a power input feature named Voltage Phase Angle (R1-PA1:VH) with values ranging from -1.7 to 0.0 did not contribute any meaningful change to the SHAP value as it remains flat at 0.0,  cases where values of Voltage Phase Angle ranging from 0.5 to 0.8, combined with the interaction of a power input feature named Current Phase Magnitude (R2-PM5:I), decreases the predicted SHAP values when compared with a case close to 1.0, where the interaction of the said Current Phase Magnitude is associated with increase in the predicted SHAP values.
\begin{figure}[htbp]
\centerline{\includegraphics[width=0.75\linewidth]{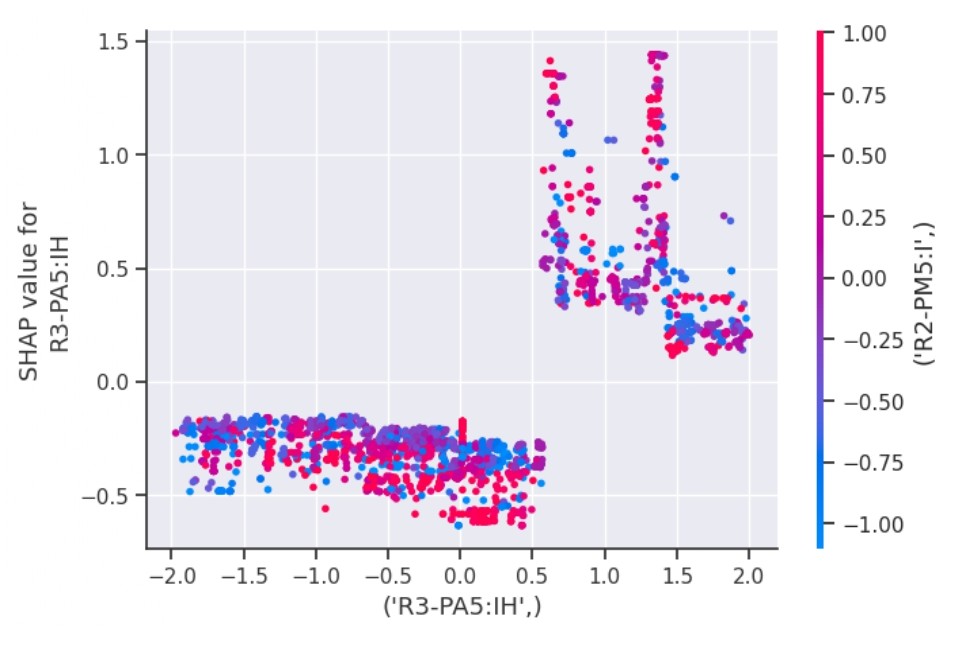}}
\renewcommand{\thefigure}{5(e)}%
\caption{Dependence Plot for R3-PA5:IH}
\label{fig}
\end{figure}

\vspace{-10pt}

In figure 5(e), we see a dependence plot showing two disjoint scenarios, one vertical, the other horizontal. From the latter, we observe cases where a power input feature named Current Phase Angle (R3-PA5:IH), with values ranging from -1.8 to 0.5, decreases the predicted SHAP values when combined with the interaction effect of an input feature named Current Phase Magnitude (R2-PM5:I); while in the former, the power input feature of Current Phase Magnitude, with values ranging from 1.7 to 2.0 produces an interaction effect that is associated with increase in the predicted SHAP values.

\begin{figure}[htbp]
\centerline{\includegraphics[width=0.75\linewidth]{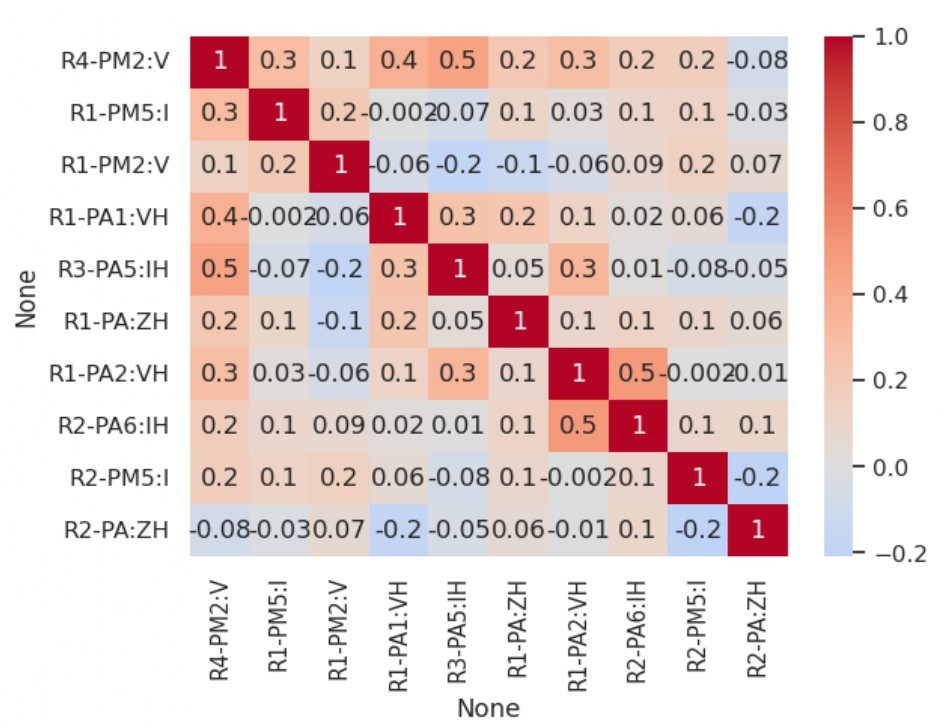}}
\renewcommand{\thefigure}{6}%
\caption{Heatmap}
\label{fig}
\end{figure}

In the figure 6 below, each square shows the correlation between the features on each axis with correlation ranges from -1 to +1. Values closer to zero means there is no linear relationship between the two features. The close to 1 the correlation is, the more positively correlated they are; meaning that, as one feature increases, so does the other; and the closer to 1 their correlation value, the stronger their relationship. Same with a correlation value closer to -1, but instead of both features increasing, one feature will decrease as the other increases. 
From the heatmap in figure 6, while the diagonals are all 1, or dark brown, denoting a perfect correlation, as the squares are correlating each feature to itself; the more positively correlated are the Voltage Phase Angle (R1-PA2:VH), Current Phase Angle (R3-PA5:IH) and Current Phase Angle (R2-PA6:IH) showing a correlation value of 0.5 each, and the least negatively correlated are the Voltage Phase Magnitude (R4-PM2:V) and Appearance Impedance Angle for Relays (R2-PA:ZH) showing a correlation value of -0.08 each. For the rest, the larger the number, and the darker the color; the higher the correlation between two features. For all intents and purposes, the plot is generally symmetrical about its diagonal. 

\subsection *{Summary of Findings}
\vspace{-10pt}
Generally, in all the three sub-datasets, the model demonstrates good performance on all classification metric as it relates to accurately identifying and classifying all the power system events. Starting with the ‘Natural and No-Events’ sub-dataset, the model recorded an accuracy score of 99 percent along with an F1score of 0.99. For the ‘Attacks and Natural Events’ sub-dataset, the model reported an accuracy score of 85 percent along with an F1score of 0.91. Also, for ‘Attack Events and No-Events’ sub-dataset, the model reported an accuracy score of 97 percent along with an F1score of 0.98. Therefore, with all the F1scores signifying good balance between precision and recall, we can then safely say that, the model can accurately identify both positive and negative cases effectively.

\vspace{-10pt}

\section{Conclusion}
\vspace{-10pt}
This paper assesses cybersecurity of smart grid exposure to cyber attacks using machine learning based approach. The study addresses the vulnerability of the sychrophasor network which can be affected not only by network cyber attacks, but also by interferences of external environment, such as weather (natural events) which in turn, cause fluctuations in power grids. As one of the common access points to infiltrate the smooth operation of power system is via a synchrophasor network, the study shows how sychrophasor network and the measurement PMU device can be attacked and manipulated, in addition to how the impacts of such an attack can be analyzed and classified through Xgboost, a machine learning model. 

\subsection *{Future Research Direction }
\vspace{-10pt}
With only 127 input features being used out of 892, future research will focus on increasing the input features, as well as scaling the dataset beyond 4,966. Additionally, to further achieve optimal performance, careful hyperparameter tuning is needed in consideration for computational intensity and large dataset. Secondly as the detection speed is critical to guarantee a timely response by the grid operators in many scenarios under attack detection, it is important to study, in future, the trade-off between the detection reliability and the detection speed. 

\end{document}